\documentclass[letterpaper, 10 pt, conference]{ieeeconf}  % Comment this line out if you need a4paper

\IEEEoverridecommandlockouts                              % This command is only needed if 
\usepackage{color}
\usepackage{amsmath}
\usepackage{amsfonts}
\usepackage{bbm}
\usepackage[dvipsnames]{xcolor}
\usepackage{subfig}
\usepackage{makecell}
\usepackage{graphicx}
\usepackage{multirow}
\usepackage{wrapfig}
\usepackage{cite}
\usepackage{makecell}
\usepackage{comment}
\usepackage{cleveref}
\usepackage{booktabs}

\usepackage[moderate,tracking=normal,charwidths=tight,paragraphs=tight,floats=normal,mathspacing=normal,mathdisplays=normal,wordspacing=normal,lists=tight]{savetrees}

\newcommand{\State}{\mathcal{S}}
\newcommand{\Action}{\mathcal{A}}
\newcommand{\Reward}{\mathcal{R}}
\newcommand{\policyparam}{{\theta}}
\newcommand{\policy}{{\pi_\theta}}
\newcommand{\expert}{{\pi_E}}
\newcommand{\E}{\mathbb{E}}
\newcommand{\Dataset}{\mathcal{D}}
\newcommand{\Discrimparam}{{\omega}}
\newcommand{\Discrim}{{D_\omega}}
\DeclareMathOperator*{\argmax}{arg\,max}
\DeclareMathOperator*{\argmin}{arg\,min}

\makeatletter % changes the catcode of @ to 11
\newcommand{\linebreakand}{%
  \end{@IEEEauthorhalign}
  \hfill\mbox{}\par
  \vspace{-1.2em}
  \mbox{}\hfill\begin{@IEEEauthorhalign}
}
\makeatother % changes the catcode of @ back to 12

\title{\LARGE \bf
Hierarchical Model-Based Imitation Learning\\for Planning in Autonomous Driving
}

\author{
  Eli Bronstein \and
  Mark Palatucci$^{1}$ \and
  Dominik Notz \and
  Brandyn White \and
  Alex Kuefler \and
  Yiren Lu \and
  Supratik Paul \and
  Payam Nikdel$^{1}$ \and
  Paul Mougin \and
  Hongge Chen \and
  Justin Fu \and
  Austin Abrams \and
  Punit Shah \and
  Evan Racah$^{1}$ \and
  Benjamin Frenkel \and
  Shimon Whiteson \and
  Dragomir Anguelov% <-this % stops a space

\thanks{All authors with Waymo Research. Contact: {\tt\small shimonw@waymo.com}}%
\thanks{$^{1}$Work performed while employed at Waymo.}%
}

\begin{document}

\maketitle
\thispagestyle{empty}
\pagestyle{empty}

%%%%%%%%%%%%%%%%%%%%%%%%%%%%%%%%%%%%%%%%%%%%%%%%%%%%%%%%%%%%%%%%%%%%%%%%%%%%%%%%
\begin{abstract}
% The difficulty and nuance of driving makes it challenging to design a rule-based system or specify a reward that matches human preferences. In contrast, imitation learning only requires a dataset of demonstration to scalably develop a planning model for autonomous vehicles.
We demonstrate the first large-scale application of model-based generative adversarial imitation learning (MGAIL) to the task of dense urban self-driving. We augment standard MGAIL using a hierarchical model to enable generalization to arbitrary goal routes, and measure performance using a closed-loop evaluation framework with simulated interactive agents. We train policies from expert trajectories collected from real vehicles driving over 100,000 miles in San Francisco, and demonstrate a steerable policy that can navigate robustly even in a zero-shot setting, generalizing to synthetic scenarios with novel goals that never occurred in real-world driving. We also demonstrate the importance of mixing closed-loop MGAIL losses with open-loop behavior cloning losses, and show our best policy approaches the performance of the expert. We evaluate our imitative model in both average and challenging scenarios, and show how it can serve as a useful prior to plan successful trajectories.

\end{abstract}

%%%%%%%%%%%%%%%%%%%%%%%%%%%%%%%%%%%%%%%%%%%%%%%%%%%%%%%%%%%%%%%%%%%%%%%%%%%%%%%%
\section{INTRODUCTION}

Driving at scale in dense urban environments remains difficult due to the complexity of interactions between large numbers of diverse actors. In these scenarios, it is difficult to apply classic motion planning methods that require defining cost functions such that the emergent behavior fully aligns with human expectations \cite{rewardMisdesign}.
This motivates imitation learning (IL), which uses expert demonstrations to learn either a cost function or a policy directly over actions \cite{Pomerleau1989-kb,Bojarski-nvidia,chauffeurnet-bansal}.

% However, imitation learning still has many difficulties that limit its practical deployment in real-world driving.
In practice, any imitation model used for motion planning needs additional safety considerations to enforce hard constraints such as collision avoidance and kinematic feasibility.
Nonetheless, studying the driving ability of an imitative model in isolation gives an indication of when and where it produces a feasible prior that could be used in an AV stack to plan successful trajectories. In these situations, motion planning could reduce to verifying trajectories from a model instead of generating them with bespoke solutions. 

A common challenge with IL is covariate shift, also known as the ``DAgger problem" \cite{ross2011reduction}. This occurs when the policy makes small errors that cause it to visit states outside of its training distribution, resulting in compounding error and divergent behavior. Intuitively, this occurs when the policy encounters unfamiliar states, and is similar to challenges in offline reinforcement learning \cite{Levine2020-rx}.
% This is when the policy visits states different from those in its training distribution, resulting in compounding error and divergent behavior. Intuitively, this occurs when the policy encounters unfamiliar states, and is similar to challenges in offline reinforcement learning \cite{Levine2020-rx}. 

State-of-the-art imitation methods like model-based generative adversarial imitation learning (MGAIL) \cite{baram2017end} address covariate shift through \emph{closed-loop training}, where dynamics are simulated and losses are backed up over the time horizon. Hence, the value of a decision depends on its long-term consequences, in contrast to open-loop behavior cloning, which treats each timestep independently. While theory predicts the importance of closed-loop training \cite{swamy2021moments}, empirical evidence in the self-driving literature is limited \cite{symphony}. This work affirms the benefit of closed-loop training on a practical, large-scale, and difficult motion planning task.

In autonomous driving, a learned motion policy must not only realistically imitate the expert, but also be goal-directed.
Such a policy can be challenging to develop due to the confounding of high-level task planning and low-level motion planning \cite{garrett2020integrated}: often a trajectory is observed from the expert, but the high-level intents or goals that affect lane choice, future route, or final destination are hidden, making it difficult to recover the causal factors that led to the observed trajectory.
One promising solution is the use of hierarchical methods, which decompose the problem into a high-level goal generation module and a low-level goal-conditioned motion policy \cite{mandelkar2020iris, Ding2019-ce}.
During training, this allows the motion policy to associate the goal with the expert's intent-driven behavior.
At inference time, this approach offers the flexibility to specify novel goals and generalize beyond the observed expert trajectories.

% Applying and evaluating MGAIL, or more broadly any imitative method, in a self-driving context is not straightforward, and involves several important design considerations that we address. One challenge is the confounding of high-level task planning and low-level motion planning \cite{garrett2020integrated}: often a trajectory is observed from the expert, but the high-level intents or goals that affect lane choice, future route, or final destination are hidden, making it difficult to recover the causal factors that led to the observed trajectory.

Ultimately, we aim to develop a policy that can safely navigate a diversity of driving situations and accomplish novel goals, including those not demonstrated by the expert. While we show that employing closed-loop training with respect to the ego vehicle's dynamics is instrumental in creating such a policy, an important question is how to properly evaluate it. Given a dataset of driving scenes with logged vehicle trajectories, evaluating an imitative policy on the same goals achieved by the expert can lead to an overly optimistic performance estimate due to spurious correlations between input features \cite{dehaan2019causal}. For example, other vehicles' logged trajectories can influence the policy to follow the expert's goal, rather than actively interacting with other actors to reach its own goal.
% V2
For this reason, it is critical to evaluate the policy's ability to follow novel goals.
This poses a challenge when simulating driving because, as the autonomous vehicle (AV) diverges from its logged trajectory to achieve a new goal, other actors' logged trajectories may become unrealistic.
To address this issue, we introduce the combination of goal generalization with \emph{closed-loop evaluation}, in which the policy attempts to reach novel goals in the presence of realistic actors that react to the AV's new actions \cite{symphony}.
% V1
% Evaluation with novel goals is a challenge in the driving environment because, as the autonomous vehicle (AV) diverges from its logged trajectory to achieve a new goal, other actors' logged trajectories may become unrealistic. Thus, to properly evaluate the policy's ability to achieve novel goals, we need \emph{closed-loop evaluation}, in which we also realistically simulate the reactions of other actors to the AV's new actions \cite{symphony}.

% V0
% Thus to truly evaluate the performance of an imitative self-driving policy, we need closed-loop simulation, with novel goals that are explicitly different from those of the expert demonstrator. This is a challenge in self-driving simulation because a novel goal invalidates the reactions of the logged trajectories of other actors. In other words, to test the performance of a novel goal, we need to realistically simulate the dynamics of both the AV \emph{and} the other actors  \cite{symphony}. 

Even with the simulation of other interactive actors, it is also important to measure performance in challenging and rare scenarios. Aggregating over a large dataset can mask the performance on difficult but uncommon situations, misrepresenting the model's ability to handle the ``long-tail" \cite{jain2021autonomy}. 

The key ingredients for closed-loop, machine-learned planner development are, now for the first time, readily available: closed-loop imitation learning with MGAIL, hierarchical goal-based policies, and realistic interactive agents.
In this work, we show how to train and evaluate such a system by demonstrating the first application of MGAIL on a large and practical self-driving task of ego vehicle motion planning for dense urban driving. Our method outperforms prior imitation approaches based on pure open-loop optimization like behavior cloning, and achieves aggregate performance similar to the expert demonstrator. We report several key design choices and experimental contributions:
% that were necessary to apply MGAIL to self-driving:

\begin{itemize}
  \item We introduce a hierarchical model that combines a high-level graph-based search with a low-level transformer-based MGAIL policy, adding an intermediate set of route features to help the model generalize and follow arbitrary goal routes.
  \item We evaluate our policy's ability to follow novel goal routes alongside simulated reactive agents in closed-loop in order to obtain more realistic estimates of zero-shot generalization and allow for interaction between the policy and other actors.
%   V0
%   \item We evaluate our policy's performance in closed-loop with novel goal routes and simulated reactive agents to obtain more realistic estimates of zero-shot generalization and allow for interaction between the policy and other actors.
 \item We run experiments on both average driving and challenging scenarios to estimate ``long-tail" performance and highlight the best opportunities for hill-climbing.
  \item We run several ablations and show that augmenting MGAIL's closed-loop adversarial losses with an open-loop behavior cloning loss leads to better performance. 
%   \item We apply state-of-the-art observation encoders based on transformers and attention mechanisms to deal with noisy perception state representation.
\end{itemize}

\section{Related Work}
\label{sec:related_work}

Imitation learning (IL) has a long history in the robotics and machine learning literature, often appearing under names such as learning from demonstration, apprenticeship learning, inverse reinforcement learning, and inverse optimal control \cite{Ng00algorithmsfor,Abbeel2004-oi, Behbahani2018-wu, Ziebart2008-ow,Finn2016-no, ratliff2006maximum}. For an overview see \cite{Argall2009-ci,imitationSurvey}. IL is also closely related to offline reinforcement learning \cite{Levine2020-rx}. Theoretical understanding of IL continues to improve with the seminal work of \cite{ross2011reduction} and more recently \cite{swamy2021moments,ortega2021shaking}. Modern approaches to IL use techniques from generative adversarial networks \cite{ho2016generative, Fu2017-tj, baram2017end, Li2017-ej} and include goal-conditioning \cite{Ding2019-ce}. 

IL has been applied to autonomous driving dating to the early success of ALVINN \cite{Pomerleau1989-kb}, and more recently \cite{chauffeurnet-bansal,Bojarski-nvidia,codevilla2018endtoend, vitelli2021safetynet,chen2019learning}. Combining expert demonstrations and reinforcement learning (RL) offers promising new approaches to scalable self-driving \cite{Kendall2018-pw, jain2021autonomy}. Despite the excitement of machine learning as a path towards large-scale deployment of AVs, many AV companies still rely heavily on classic search-based planning and trajectory optimization. For a survey of classic approaches, see \cite{Paden2016-ix}. 

While motion forecasting models \cite{varadarajan2021multipath} have had a long history in AV stacks to predict other agent behavior, recent work has applied these models to the ego agent to predict feasible trajectories for direct planning, and can be viewed in the context of open-loop imitation \cite{rhinehart2019precog, rhinehart2019deep, liu2021deep, casas2021mp3, ettinger2021large, zeng2021endtoend,ngiam2021scene}. 

Closed-loop simulation continues to advance, both for the purposes of evaluating driving performance through realistic world models \cite{symphony,suo2021trafficsim,nuplan,zhou2020smarts, Dosovitskiy2017-ye,okelly2019scalable}, and to train driving policies that could transfer to the real world \cite{pan2017virtual,muller2018driving}.

The work most similar to ours is \cite{symphony}, which applies model-based imitation and parallel beam search to train simulated agents for testing an AV. By contrast, we focus on the ego AV motion planning problem, where following arbitrary goal routes is critical.
We
% use modern observation encoders based on transformers and 
avoid beam search as done by \cite{symphony}, which requires future information from reference trajectories not available in the context of ego agent motion planning \cite{liu2021multimodal,jaegle2021perceiver}. We also focus on dense urban driving.

\section{Background}
\label{sec:background}

% In this section, we review our problem formulation, notation, and cover background information on GAN-based imitation learning.

\subsection{Markov Decision Processes}
\label{sec:mdp}

We formulate the planning problem as a Markov Decision Process, defined by the tuple $(\State, \Action, \mathcal{T}, \Reward, \gamma, p_0)$. $\State$ represents the state space, which encapsulates quantities such as the positions and velocities of cars, road features, and dynamic objects within the scenario. $\Action$ represents the action space, $\mathcal{R}$ the reward function, $\gamma$ the discount factor, $\mathcal{T}$ the transition distribution, and $p_0$ the initial state distribution. A policy, or decision making agent, with parameters $\policyparam$ is denoted by $\policy$. 
A typical goal is to find a policy that maximizes the expected return, $\E_{\tau \sim \policy}\left[ \sum_{t=0}^T \gamma^t r(s_t, a_t)\right]$, where $\tau = \{s_0, a_0, \dots, s_T, a_T\}$ denotes a trajectory.

Many classes of techniques have been proposed to find good policies. Methods such as reinforcement learning and optimal control can be used to directly find a strong policy when the reward function is available~\cite{deisenroth2013survey}. However, it can be difficult to design a suitable reward function in many application areas, including autonomous driving \cite{rewardMisdesign}. An alternative approach is imitation learning (IL), in which we assume access to a dataset of trajectories $\Dataset = \{\tau_i\}_{i=0}^{N}$~\cite{Argall2009-ci} demonstrated by an expert, denoted as $\expert$. The objective is then to train a policy to imitate the demonstrations.

\subsection{Model-based Generative Adversarial Imitation Learning}
\label{sec:mgail}

The naive approach to imitation learning is to perform supervised learning directly on the expert demonstrations, an approach commonly known as behavior cloning (BC) \cite{michie1990cognitive,Pomerleau1989-kb}:
\begin{align*}
    \argmax_{\policyparam}\ \E_{s, a \sim \pi_{E}}[\log\ \policy (a| s)].
\end{align*}

However, a policy trained in this way is susceptible to covariate shift at test time, resulting in accumulating error.
% However, when a policy trained in this way is deployed at test time, small errors can cause the policy to visit states outside the training distribution, leading the policy to make more mistakes.
This means that most BC methods suffer from a quadratic worst-case error scaling with respect to the length of the episode~\cite{ross2011reduction}.

Generative adversarial imitation learning (GAIL)~\cite{ho2016generative} is an alternative that treats imitation learning as an adversarial game. GAIL samples trajectories using the current policy, trains a discriminator to classify these trajectories against the demonstrations (where the demonstrations are labeled as 1 and sampled trajectories as 0), and then optimizes the policy to make its trajectories indistinguishable from the demonstrations. This can be formalized as finding a Nash equilibrium in the following minimax game between the policy $\policy$ and the discriminator $\Discrim$ (parameterized by $\Discrimparam$):
\begin{align*}
    \argmax_\policyparam\ \argmin_\Discrimparam\  &\E_{s, a \sim \policy}[\log\ \Discrim(s, a)] + \\
    &\E_{s, a \sim \expert}[\log(1 - \Discrim(s, a))].
\end{align*}

Theoretically, rather than minimizing the error in the conditional distribution $\policy(a|s)$ as in BC, GAIL minimizes the gap in the occupancy measures (the joint distribution of states and actions $p(s, a)$) between the policy and the expert. This allows GAIL to achieve linear error scaling~\cite{swamy2021moments} in the time horizon, greatly reducing the effect of compounding error at the cost of requiring online rollouts through a simulator.
% Note that MGAIL may not satisfy this linear error scaling property, since it optimizes the policy open-loop.

Model-based generative adversarial imitation learning (MGAIL)~\cite{baram2017end} utilizes a differentiable dynamics model to improve training.
GAIL requires the use of high-variance policy gradient estimates or other zero-order optimization methods to optimize the policy because its training objective is not differentiable due to sampling through an unknown transition model. With a differentiable model, gradient estimators such as the reparametrization trick~\cite{xu2019variance} can be used to reduce the variance of the policy update. We leverage the fact that our driving simulator is differentiable, as described in section \ref{sec:deltaap}, in order to apply MGAIL to our method.

%\subsection{On Policy Reward Matching}
%JF: This is mostly covered in the section above.

\section{Method}

In this section, we first provide an overview of our planning agent's system architecture, including the high-level route generation module and low-level continuous motion policy.
We then explain how to evaluate the policy in a data-driven, closed-loop simulation with realistic, interactive agents in order to test generalization to novel goals.
Finally, we describe an approach to scalably test the policy's robustness to challenging and rare driving situations.

\subsection{Route Generation}

\begin{figure}[ht]
\centering
\includegraphics[width=0.95\columnwidth]{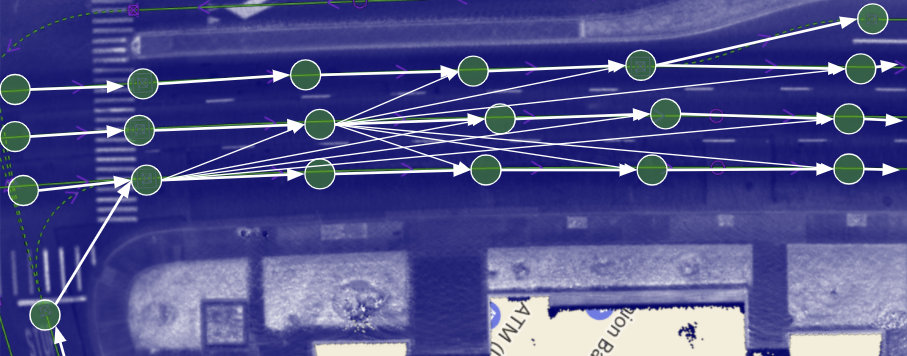}
\caption{The routing graph built on a lane-level map. Nodes are fixed lane waypoints, and edges are maneuvers between them (not all lane changes are shown for clarity).}
% \caption{The routing graph built on a lane-level map. Nodes are fixed waypoints in each lane, and edges are maneuvers between them (stay in lane, turn, fork, and lane change edges are shown; not all lane changes shown for clarity).}
\label{fig:lane_waypoint_graph}
%original drawing file: https://docs.google.com/presentation/d/1NrpuGVp353JskwbeyiCrpY7jEulbnR6sjMimHORa35c/edit#slide=id.g1df0e8e398_0_147
\end{figure}

In the high-level module, we generate routes across waypoints on a premapped lane-level roadgraph, where nodes are discretized lane waypoints, and edges are maneuvers between them: stay in the lane, take a turn/merge/fork, or execute a lane change (Figure~\ref{fig:lane_waypoint_graph}).

Our primary routing cost is proportional to historical traversal time for each maneuver, so as to minimize ETA.
%% Alternative to consider:
% We use a combination of learned routing costs (e.g. a model for historical traversal time) and hand-tuned costs (e.g. a flat cost per turn to encourage more direct routes).
Using a bidirectional A* search with the ALT heuristic \cite{goldberg2004heuristic}, we generate one or more routes, depending on the roadgraph.
During data collection, the expert is tasked with following the lowest cost route.
For training, we condition the low-level motion policy on this same route.
However, for evaluation, other routes may be chosen to test the policy's generalization capabilities.

% Additional router details:
% We also use a set of dynamic costs dependent on perceived objects within the car's field of view (e.g. avoid nearby double-parked vehicles and active construction zones).
% This search maintains its optimal path-to-goal over time, similar to Reverse Resumable A*\cite{silver2005resumable}, so that we can re-search 10 times per second to produce the optimal path as the car moves through the world.

\subsection{Observation Encoding}

Transformer architectures \cite{vaswani2017attention} have been shown to be effective for modeling high-dimensional sequences. We use stacked transformer-based models~\cite{liu2021multimodal,mercat2020multi} to encode multimodal observations for the MGAIL policy and discriminator. 

We divide the input features into five groups: the AV’s own trajectories (e.g., position, heading), other context objects’ trajectories (e.g., other vehicles and pedestrians), roadgraph points (positions and types of points sampled from continuous road network features), traffic light signals, and the goal route generated by the high-level module (represented as a sequence of points).

\begin{figure*}[ht]
\vspace{2mm}
\centering
\includegraphics[width=2\columnwidth]{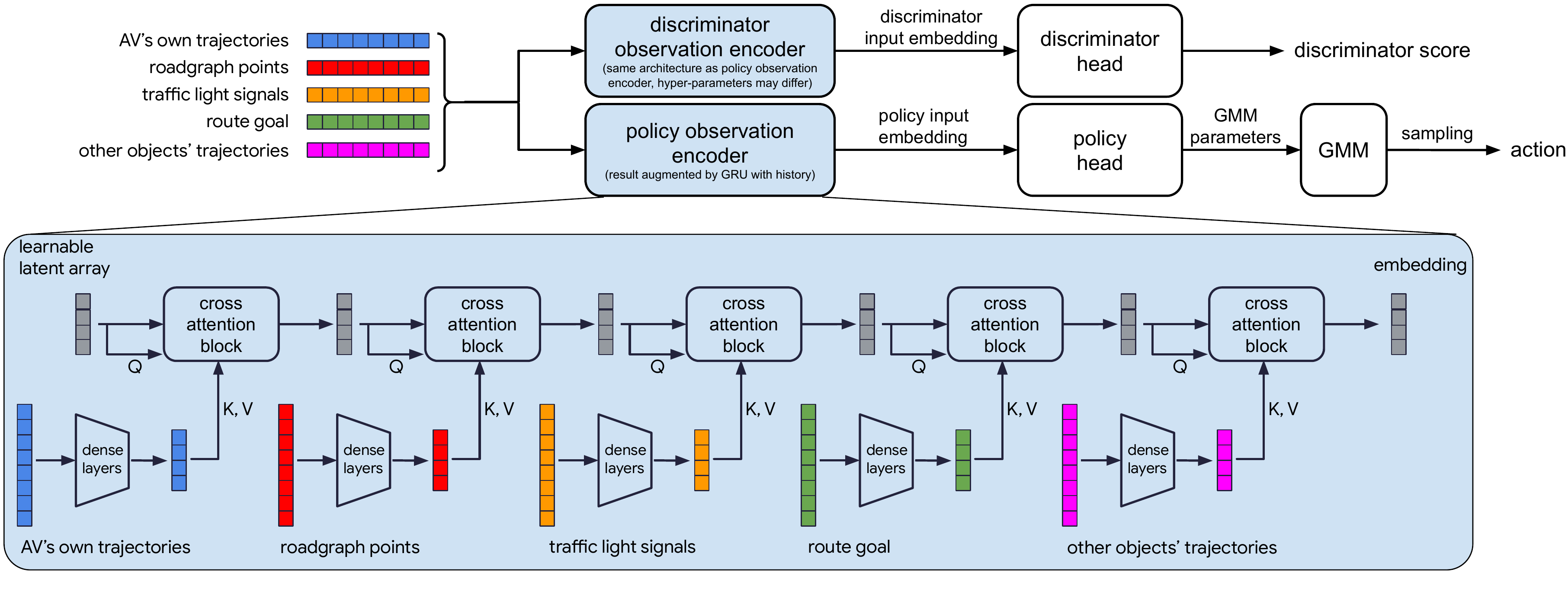}
\caption{System diagram of the MGAIL planning agent and its transformer-based observation encoder. ``K", ``V", and ``Q" represent the keys, values, and queries, respectively.}
%\vspace{-5mm} 
\label{fig:perceiver_encoder}
%original drawing file, can be edited directly: https://docs.google.com/drawings/d/1bX4Q2lMeONczGaJJWEg03hLmybIs9oWyaFrkloe_qTE/edit?usp=sharing&resourcekey=0-3yVz4Fil4NGjq-fRqcZUKw
\end{figure*}

The computational complexity and memory usage of conventional transformer models' cross-attention stages grow quadratically with the features' input dimension.
Instead, we use cross-attention with linear computational complexity in terms of the input dimension.
Inspired by Set-Transformer~\cite{lee2019set} and Perceiver~\cite{jaegle2021perceiver} approaches, we use learned arrays as queries to fuse the multimodal inputs.
This design is flexible since we can fuse the inputs in any order, which is chosen as a hyperparameter based on empirical results.
As shown in Figure~\ref{fig:perceiver_encoder}, we first concatenate features in each group and encode them separately using dense layers (size (32, 32), L2 regularization with weight 0.01, and layer normalization~\cite{ba2016layer}).
The dimensions of each group's output embeddings are the same.
The model is a stack of cross-attention blocks that iteratively attend to each group of features with a fixed-dimensional learnable latent bottleneck (sizes 128 and 64 for the policy and discriminator)
% and optionally multiple heads (four for the policy, one for the discriminator)
.
Each cross-attention block has a residual attention layer followed by a dense layer.
The output policy and discriminator observation embeddings are then used as inputs to the MGAIL policy and discriminator heads, respectively.
For the policy head only, we employ a gated recurrent unit (GRU)~\cite{cho2014properties} to integrate the output policy observation embeddings over time.

\subsection{Delta Actions Model}
\label{sec:deltaap}

% In most simulated control tasks, e.g., MuJoCo \cite{todorov2012mujoco}, the transition dynamics model is provided as part of the environment and is considered fixed.
% By contrast, in autonomous driving simulation, the overall dynamics of the environment depends on all the agents \cite{zhou2020smarts}.
% In our experimental setting, the surrounding objects could be either following the logged trajectories or be replaced by interactive agents; we have the flexibility to choose or design the low-level single agent dynamics model.

While the trajectories of other vehicles observed during data collection are given, we must choose a dynamics model for the AV.
We utilize the delta actions model, which uses offsets to the current state $s$ as actions $a$ to compute the next state $s' = s + a$.
Demonstrated actions can be easily obtained by differentiating consecutive demonstrated trajectory states: $a = s' - s$.
This model is fully differentiable but is not learned from data.
Unlike traditional vehicle dynamics models, this model does not enforce constraints on the behavior space.
This characteristic makes it flexible enough to cover a diverse set of driving behaviors, and it could even be used for non-vehicle agents, such as pedestrians.
In practice, an AV stack would include a local controller that attempts to realize the next state generated by the dynamics model.

For the policy, we use a Gaussian Mixture Model (GMM) to parameterize the delta actions, where the GMM parameters are predicted by a neural network.
The distribution over actions $a$ given a state $s$ is given by:
\begin{align*}
\pi_\theta(a|s) =& \sum_{i=1}^K \phi_\theta^i (s) \mathcal{N} (a | \mu_\theta^i (s), \Sigma_\theta^i (s) ),
% N (a | \mu_i (x), \sigma_i (x) ) =& \frac{1}{\sigma_i (x) \sqrt{2\pi}} \exp{- \frac{(a - \mu_i(x))^2}{2\sigma_i(x) ^ 2} } \\
% \sum_i^K \phi_i =& 1
\end{align*}
where $\mathcal{N}(\cdot)$ denotes a Gaussian distribution, $K$ (set to 8) is the number of Gaussians, and $\phi_\theta^i$, $\mu_\theta^i$, and $\Sigma_\theta^i$ for $i \in \{1, \ldots, K\}$ are the weights, means, and covariances predicted by the policy network given a state $s$ (see Figure~\ref{fig:perceiver_encoder}). During inference, actions are sampled from the predicted GMM.

A common alternative approach is to directly predict the action values and to minimize the mean squared error (MSE) between the predicted and demonstrated actions. This approach is equivalent to maximizing the log likelihood of the demonstrated action assuming it is sampled from a unimodal Gaussian distribution. However, in our setting multiple actions may be appropriate for a given state, which motivates using a GMM to represent different modes.

% \subsection{\revision{Training}}
\subsection{Training Losses}
\label{section:losses}

For the low-level continuous motion policy, we combine MGAIL with behavior cloning (BC).
The total loss consists of three main components: the discriminator loss $\mathcal{L_{D}}$, the MGAIL policy loss $\mathcal{L_{P}}$, and the BC loss $\mathcal{L_{BC}}$:
\begin{align}
\mathcal{L} = \mathcal{\lambda_{D}} \mathcal{L_{D}} + \mathcal{\lambda_{P}} \mathcal{L_{P}} + \mathcal{\lambda_{BC}} \mathcal{L_{BC}},
\end{align}
where $\mathcal{\lambda_{D}}$, $\mathcal{\lambda_{P}}$ and $\mathcal{\lambda_{BC}}$ are the coefficients for each loss component. To assess the individual impacts of the MGAIL and BC policy losses, we can adjust the contribution of loss components by reducing or zeroing their coefficients. The BC loss is:
% \noindent The BC loss is given by:
\begin{align}
    \mathcal{L_{BC}}=&- \E_{s,a\sim\pi_{E}}[\log\ \policy (a| s)].
\end{align}
The MGAIL discriminator loss is:
\begin{align}
    \mathcal{L_{D}}=& \E_{s\sim\policy}[\log\ \Discrim(s)] + \E_{s\sim\expert}[\log(1 - \Discrim(s))].
\end{align}
Here we only provide the state $s$ to the discriminator as input similar to \cite{torabi2018generative}. When performing backpropagation, only the discriminator parameters $\omega$ are updated. The MGAIL policy loss is:
\begin{align}
\mathcal{L_{P}}=& -\E_{s\sim\policy}[\log\ \Discrim(s)].
\end{align}
During backpropagation, only the policy parameters $\theta$ are updated for this loss term. Since our dynamics model is differentiable, the gradients are propagated back in time and influence the actions of the policy in earlier timesteps \cite{baram2017end}. We use the reparametrization trick~\cite{xu2019variance} to differentiate through the stochastic GMM policy. We minimize these losses using Adam~\cite{kingma2014adam} (default parameters), a learning rate of 0.0003, and gradient clipping (max L2 norm of 10).

\subsection{Closed-Loop Evaluation with Interactive Agents}
\label{sec:agent_reactions}

We evaluate the policy by having it control the AV in driving situations recorded during data collection.
To test the policy's ability to reproduce expert behavior, we can condition it on the expert's goal route, while having other road users (RUs) follow their logged trajectories.
However, an ideal AV policy should be able to generalize to novel routes that differ from those followed by the expert.
In this situation, if other RUs simply follow their logged trajectories, the simulation is likely to become unrealistic due to the lack of interaction between the AV and other RUs.
As the AV diverges from its logged trajectory, RUs should be able to diverge in response.
If they do not, following a novel route may be infeasible without causing a traffic conflict.

We therefore study the performance of our policy's zero-shot generalization to novel routes in the presence of an interactive Symphony agent \cite{symphony} that governs the behavior of RUs near the AV.
The Symphony agent is trained separately using a transformer observation encoder and delta actions model using a combination of MGAIL and BC losses, similar to the AV policy.
However, it is not goal-conditioned because its objective is simply to drive realistically rather than achieve a specific goal.
% Unlike the AV policy, we train the Symphony agent using RU trajectories to imitate human driving behavior.
This allows us to evaluate the goal-directed planning agent in a realistic, closed-loop simulation that more accurately reproduces the complex interactions of real-world driving.

\subsection{Challenge Classifier}
\label{sec:risk_classifier}

While evaluating on randomly sampled segments yields an unbiased estimate of the policy's performance, it is not very informative of how it handles difficult scenarios.
Since driving is a safety-critical task, it is crucial to understand our model's failure modes in rare and challenging settings.

With this in mind, we develop a ML classifier to score segments on their potential for producing a realistic collision or ``close call" when simulated with a planning agent.
The classifier takes as input the segment and outputs a score in $[0, 1]$ with higher values indicating higher likelihood of a simulated collision or close call.
We collect training data by simulating a planning agent on logged data and asking human labelers to identify when a realistic collision or close call occurred, if at all.
Timesteps at a $\pm 3$ second offset are positive examples (label $1$) and the remaining timesteps are negative examples (label $0$).
Since collisions and close calls are rare, during training we upsample segments that contain positive examples.
We use a fully connected multi-layer perceptron with (1024, 1024, 512, 32) hidden layers.
To prevent train-test leakage, we train this classifier on a dataset collected at least 3 months prior to those used for the experiments below.
This dataset consists of 145k trajectories, 5.6k of which were deemed to have realistic collisions or close calls by human labelers.
The remaining negative trajectories do not require human labels.
Using the classifier's scores for the entire test set, we can evaluate the policy on subsets of varying challenge levels.

\section{Experiments}
\label{sec:experiments}

\subsection{Dataset}    
We collected a large sample of 10 million expert trajectories, each a segment of 10 seconds, from a fleet of our vehicles operating in San Francisco.
During data collection, the high-level module generates one or more routes toward a destination, and the expert attempts to follow the best route.
These trajectories represent over 100,000 miles of expert driving, which we split into 80\% \emph{train}, 10\% \emph{validation}, and 10\% \emph{test} sets. Each set contains disjoint \emph{runs}, meaning that trajectories from the same vehicle operating on the same day exist only within one set. The trajectories are sampled at 15 Hz, and represent ego vehicle state (e.g., pose) and exteroceptive state from a robotic perception system (e.g., other actors' bounding boxes, traffic light state). These trajectories are aligned with the coordinate frame of a high-definition map \cite{ettinger2021large}. 

\subsection{Metrics}

We measure driving performance using three key metrics, each of which reports a success/failure indicator for the segment. Specifically, we compute each metric per timestep, and a failure at any timestep indicates a failure for the segment.
\begin{itemize}
  \item \emph{Road-Route Failure}: If the trajectory deviates from the goal road-route, which is a traversal through a road network that allows the vehicle to be in any valid lane along the route, similar to a Google Maps style route. 
  \item \emph{Collision}: If the bounding box of the ego vehicle intersects with a bounding box of another object. 
  \item \emph{Off-road}: If the bounding box of the ego vehicle deviates from the drivable surface according to the map.
%  \item \emph{Red-Light Violation} (RLV):  If the bounding box the vehicle intersects the bounding box of an %intersection when the traffic light controlling its lane is red. 
\end{itemize}
We combine these metrics to generate an overall success metric for a segment (i.e., a segment is marked successful if it follows the road-route without any collisions or off-road driving).
We report the average of each metric, and the overall success rate for the given test set.
We also report the \emph{Route Progress Ratio}, which is the ratio of the distance traveled along the route by the policy vs. the expert, given that both followed the route.
We use this metric to ensure that the policy makes comparable progress to the expert (higher is better).
For each experiment, we specify the number of test segments used and report results with 95\% confidence sets using the normal approximation.
% In each experiment detailed below, we specify the exact number of test segments used. All results are reported along with 95\% confidence sets using the normal approximation.
Finally, in the spirit of Goodhart's law, none of these metrics are visible to the model in our experiments, although loss or reward shaping for safety and to enforce constraints is a natural area for future work \cite{vitelli2021safetynet,amodei2016concrete}.

We simulate the policy trajectories, play back the expert trajectories, and compute metrics in our simulator using bounding box approximations for vehicle geometry.
References to the expert performance are labeled in the results tables as \emph{Playback}.

\subsection{Results}

In this section, we present results of several model variants and ablate key modeling and evaluation decisions, using BC and vanilla MGAIL as baselines.
We assess the utility of goal route conditioning, compare the performance on logged routes and novel routes to evaluate the policy's generalization capability, and highlight the importance of closed-loop evaluation with interactive agents.
We also demonstrate the policy's performance on more challenging test subsets, as well as the best mixture of MGAIL and BC training losses.

\subsubsection{\textbf{Can the model safely follow logged routes? What is the impact of route conditioning?}}

In Table \ref{tab:1} we compare six model variants to the expert on the \emph{Unbiased Test Set} - an unbiased sample of dense urban driving consisting of 82,198 segments.
We initialize the simulator to the initial state of a scene and compute the route once at the beginning of the segment, keeping it fixed for the duration of the rollout. In this setup, we test the model's ability to successfully follow the \emph{logged route}, which is provided as a conditioning feature.
In other words, we ask the model to drive the same route that the expert attempted to follow during the test segment, which is the best route produced by the high-level route generation module.
In these experiments, the model only controls the AV, while other vehicles follow their logged trajectories.

``BC" indicates behavior cloning loss ($\mathcal{L_{BC}}$) only, ``MGAIL" means adversarial losses ($\mathcal{L_{D}}$ and $\mathcal{L_{P}}$) only, and ``MGAIL + BC" indicates a mixture of adversarial and BC losses, where the relative weight of each adversarial loss to the BC loss is 2:1. The weights of both MGAIL adversarial losses are equal.

Without the route conditioning features all model variants struggle to follow the route, even if their overall collision and off-road rates are low. The route conditioning features significantly improve performance of all models, with our best MGAIL + BC variant achieving 99.6\% of the expert's success rate. The route features also decrease off-road violations for all variants, and they reduce collisions for the MGAIL and MGAIL + BC variants. We also observe that for the segments where both the policy and the expert followed the road-route, each policy variant makes comparable or slightly more progress along the route compared to the expert.

\begin{table*}[thpb]
\vspace{2mm}
\centering
\caption{Logged Route on the \emph{Unbiased Test Set}.}
\label{tab:1}
\renewcommand{\arraystretch}{1.3}
% \hspace{-8mm}
\resizebox{.7\textwidth}{!}{
\begin{tabular}{l|cccc|c}
\Xhline{2\arrayrulewidth}
         \multirow{2}{*}{Method}
         & \begin{tabular}[c]{@{}c@{}}Success\\ rate (\%)\end{tabular} 
         & \begin{tabular}[c]{@{}c@{}}Route Failure\\ rate (\%)\end{tabular}  
         & \begin{tabular}[c]{@{}c@{}}Collision\\ rate (\%)\end{tabular} 
         & \begin{tabular}[c]{@{}c@{}}Off-road\\ rate (\%)\end{tabular}
         & \begin{tabular}[c]{@{}c@{}}Route Progress\\ratio (\%)\end{tabular}\\
\Xhline{2\arrayrulewidth}
Playback & 98.62{\scriptsize $\pm$0.08} & 1.07{\scriptsize $\pm$0.07} & 0.05{\scriptsize $\pm$0.02} & 0.26{\scriptsize $\pm$0.03} & 100.00{\scriptsize $\pm$0.00}\\
\hline
BC & 86.07{\scriptsize $\pm$0.24} & 9.63{\scriptsize $\pm$0.20} & 4.53{\scriptsize $\pm$0.14} & 2.21{\scriptsize $\pm$0.10} & 105.59{\scriptsize $\pm$0.40}\\
BC + Route & 94.18{\scriptsize $\pm$0.16} & \bf 0.69{\scriptsize $\pm$0.06} & 4.60{\scriptsize $\pm$0.14} & 0.75{\scriptsize $\pm$0.06} & 98.10{\scriptsize $\pm$0.33}\\
MGAIL & 88.90{\scriptsize $\pm$0.21} & 9.73{\scriptsize $\pm$0.20} & 1.28{\scriptsize $\pm$0.08} & 1.00{\scriptsize $\pm$0.07} & 101.22{\scriptsize $\pm$0.32}\\
MGAIL + Route & 97.45{\scriptsize $\pm$0.11} & 0.74{\scriptsize $\pm$0.06} & 1.20{\scriptsize $\pm$0.07} & 0.77{\scriptsize $\pm$0.06} & 100.85{\scriptsize $\pm$0.29}\\
MGAIL + BC & 89.84{\scriptsize $\pm$0.21} & 8.93{\scriptsize $\pm$0.19} & 1.25{\scriptsize $\pm$0.08} & 0.73{\scriptsize $\pm$0.06} & 105.58{\scriptsize $\pm$0.36}\\
MGAIL + BC + Route & \bf 98.22{\scriptsize $\pm$0.09} & \bf 0.69{\scriptsize $\pm$0.06} & \bf 0.77{\scriptsize $\pm$0.06} & \bf 0.37{\scriptsize $\pm$0.04} & 105.30{\scriptsize $\pm$0.32}\\
\Xhline{2\arrayrulewidth}
\end{tabular}}
\end{table*}

\subsubsection{\textbf{Can the model generalize to novel routes?}}

One danger of conditioning on logged routes is spurious correlation, where the model may not properly learn the causal relationship of the route. In essence, giving the model the same goal route as the expert overestimates its ability to generalize, since other features may correlate with the original route.

To test the model's generalization to arbitrary routes, we created the \emph{Route Generalization Test Set} - a subset of the \emph{Unbiased Test Set} with 18,593 segments where the high-level route generation module produced multiple, distinct routes.
For example, one route may go straight through an intersection while another route may be a right turn.
This subset is more challenging than the entire test set because segments with multiple routes are likely to have more complex road networks and potentially more interactions with other vehicles.
In Table \ref{tab:2}, we show the policy's ability to follow \emph{novel routes}, where we select a goal route that differs from the one the expert attempted to follow (i.e., the \emph{logged route}).
In both the logged and novel route sections of the table, the initial conditions of the segments are identical.

As in our previous experiments, the route conditioned features are critical to overall success and route following. For the model variants without these features, performance drops considerably compared to following logged routes. This shows the danger of correlated features (e.g., other vehicles' trajectories), where the model tends to follow the logged trajectory despite not observing it.

We notice a generalization gap for the route conditioned variants, whose overall success rate drops by as much as 12\% compared to when following the logged route.
However, the performance of the route conditioned MGAIL variants drops less than that of BC, suggesting that MGAIL helps with generalization.
We also see that MGAIL + BC continues to perform the best, but on novel routes the value of adding the BC loss is within the margin of error of MGAIL's purely adversarial losses, in terms of the route failure and collision metrics.

As none of our metrics are included in the training losses, these results suggest that purely associative feature conditioning may be insufficient to achieve robust route generalization necessary for driving that is on par with the expert's performance. Constrained optimization and reward shaping may be required, or refinement through reinforcement learning, which we leave for future work.  

We also see that on novel routes, the route-conditioned variants have higher collision and off-road rates compared to logged routes.
This is unsurprising, since the other vehicles simply follow their logged trajectories and do not react realistically when the policy attempts to follow the novel route, making it challenging or infeasible for it to do so.
For the same reason, collision and off-road rates rise when the model is route-conditioned, compared to its unconditioned counterpart.
The unconditioned variants do not explicitly attempt to follow the novel route since they do not observe it, so they are more likely to drive in a generally realistic way that does not require other actors to be interactive.

\begin{table*}[thpb]
\vspace{2mm}
\caption{Logged Route vs.\ Novel Route on the \emph{Route Generalization Test Set}.}
\label{tab:2}
\renewcommand{\arraystretch}{1.2}
% \hspace{-8mm}
\resizebox{\textwidth}{!}{
\begin{tabular}{l|cccc|cccc}
\Xhline{2\arrayrulewidth}
         \multirow{3}{*}{Method} 
         & \multicolumn{4}{c|}{Logged Route} & \multicolumn{4}{c}{Novel Route} \\ \cline{2-9}  
         & \begin{tabular}[c]{@{}c@{}}Success\\ rate(\%)\end{tabular} 
         & \begin{tabular}[c]{@{}c@{}}Route Failure\\ rate(\%)\end{tabular}  
         & \begin{tabular}[c]{@{}c@{}}Collision\\ rate (\%)\end{tabular} 
         & \begin{tabular}[c]{@{}c@{}}Off-road\\ rate (\%)\end{tabular}
         & \begin{tabular}[c]{@{}c@{}}Success\\ rate(\%)\end{tabular} 
         & \begin{tabular}[c]{@{}c@{}}Route Failure\\ rate(\%)\end{tabular}  
         & \begin{tabular}[c]{@{}c@{}}Collision\\ rate (\%)\end{tabular} 
         & \begin{tabular}[c]{@{}c@{}}Off-road\\ rate (\%)\end{tabular} \\
\Xhline{2\arrayrulewidth}
Playback & 95.73{\scriptsize $\pm$0.29} & 3.97{\scriptsize $\pm$0.28} & 0.01{\scriptsize $\pm$0.01} & 0.31{\scriptsize $\pm$0.08} & - & - & - & -\\
\hline
BC & 68.97{\scriptsize $\pm$0.66} & 26.63{\scriptsize $\pm$0.64} & 5.62{\scriptsize $\pm$0.33} & 3.51{\scriptsize $\pm$0.26} & 20.22{\scriptsize $\pm$0.58} & 76.67{\scriptsize $\pm$0.61} & 5.72{\scriptsize $\pm$0.33} & 3.77{\scriptsize $\pm$0.27}\\
BC + Route & 92.22{\scriptsize $\pm$0.38} & 1.90{\scriptsize $\pm$0.20} & 4.94{\scriptsize $\pm$0.31} & 1.30{\scriptsize $\pm$0.16} & 80.34{\scriptsize $\pm$0.57} & 7.82{\scriptsize $\pm$0.39} & 11.09{\scriptsize $\pm$0.45} & 4.87{\scriptsize $\pm$0.31}\\
MGAIL & 69.82{\scriptsize $\pm$0.66} & 28.90{\scriptsize $\pm$0.65} & 1.69{\scriptsize $\pm$0.19} & 1.95{\scriptsize $\pm$0.20} & 25.08{\scriptsize $\pm$0.62} & 73.78{\scriptsize $\pm$0.63} & \bf 1.63{\scriptsize $\pm$0.18} & 1.97{\scriptsize $\pm$0.20}\\
MGAIL + Route & 95.91{\scriptsize $\pm$0.28} & \bf 1.86{\scriptsize $\pm$0.19} & 1.31{\scriptsize $\pm$0.16} & 1.22{\scriptsize $\pm$0.16} & 88.11{\scriptsize $\pm$0.47} & 5.80{\scriptsize $\pm$0.34} & 4.78{\scriptsize $\pm$0.31} & 3.16{\scriptsize $\pm$0.25}\\
MGAIL + BC & 71.97{\scriptsize $\pm$0.65} & 26.96{\scriptsize $\pm$0.64} & 1.72{\scriptsize $\pm$0.19} & 1.37{\scriptsize $\pm$0.17} & 20.27{\scriptsize $\pm$0.58} & 78.52{\scriptsize $\pm$0.59} & 1.69{\scriptsize $\pm$0.19} & \bf 1.32{\scriptsize $\pm$0.16}\\
MGAIL + BC + Route & \bf 97.05{\scriptsize $\pm$0.24} & 1.90{\scriptsize $\pm$0.20} & \bf 0.67{\scriptsize $\pm$0.12} & \bf 0.48{\scriptsize $\pm$0.10} & \bf 89.65{\scriptsize $\pm$0.44} & \bf 5.67{\scriptsize $\pm$0.33} & 4.41{\scriptsize $\pm$0.30} & 1.60{\scriptsize $\pm$0.18}\\
\Xhline{2\arrayrulewidth}
\end{tabular}}
\end{table*}

\subsubsection{\textbf{What is the impact of closed-loop evaluation with interactive agents?}}

In order to preserve simulation realism while the AV deviates from its logged trajectory, we introduce interactive agents to represent other RUs (section \ref{sec:agent_reactions}).
We create the \emph{Interactive Agent Test Set} by assigning the top 8 RUs to the interactive agent at the beginning of each segment based on their proximity to the AV's logged trajectory.
We only replace RUs that were fully observed by the AV's perception system during data collection and reject scenarios with fewer than 8 fully observed RUs, yielding 4,406 segments.
Due to the constraint that 8 objects must be fully observed, this dataset overrepresents busy or congested scenes. 
These scenarios may still contain partially observed RUs, but they follow their logged trajectories.%, rather than being controlled by the interactive agent.

Table \ref{tab:3} shows the zero-shot performance on novel routes with reactive agents absent and present. Because of the rejection sampling used to obtain denser scenes, this problem setting is more difficult than the \emph{Unbiased Test Set} presented in table \ref{tab:2}. However, for a fixed AV policy, including reactive agents reduces the collision rate and route failure rate, and increases the success rate. The reduction in collision rate may be explained by reactive agents avoiding conflicts with the AV. However, the fact that the AV policy improves on the route failure metrics indicates that a significant number of failures were due to routes being infeasible without initiating an interaction with another RU.
In some cases, the reactive agents may be responsible for collisions with the AV, but it is challenging to determine which agent is at fault.
There is still a gap between logged route performance and zero shot generalization to novel routes, even with reactive agents. Performance may be improved by retraining the AV policy alongside reactive agents, but we leave these experiments for future work.

% \commentmp{Agent reactions with just novel routes. Call out why different dataset (and more challenging).
% Table: BC+Route, MGAIL-D+Route, MGAIL-D+BC+Route, (5 cols for novel routes no agents, 5 cols for novel routes with agents). 3x10 table.}

% Conclusions:
% 1) harder dataset because scenes more dense. All of the metrics are worse when comparing this dataset without reactive agents to the other full dataset without reactive agents on novel routes.
% 2) agent reactions improve overall success rate, route rate, and collsion rate.
% 3) future work necessary... There's still a gap between novel routes with reactive agents and logged routes without reactive agents.

\begin{table*}[thpb]
\caption{Novel Route Without vs.\ With the Interactive Agent on the \emph{Interactive Agent Test Set}.}
\label{tab:3}
\renewcommand{\arraystretch}{1.2}
% \hspace{-8mm}
\resizebox{\textwidth}{!}{
\begin{tabular}{l|cccc|cccc}
\Xhline{2\arrayrulewidth}
         \multirow{3}{*}{Method} 
         & \multicolumn{4}{c|}{Without Reactive Agent} & \multicolumn{4}{c}{With Reactive Agent} \\ \cline{2-9}  
         & \begin{tabular}[c]{@{}c@{}}Success\\ rate (\%)\end{tabular} 
         & \begin{tabular}[c]{@{}c@{}}Route Failure\\ rate (\%)\end{tabular}  
         & \begin{tabular}[c]{@{}c@{}}Collision\\ rate (\%)\end{tabular} 
         & \begin{tabular}[c]{@{}c@{}}Off-road\\ rate (\%)\end{tabular}
         & \begin{tabular}[c]{@{}c@{}}Success\\ rate (\%)\end{tabular} 
         & \begin{tabular}[c]{@{}c@{}}Route Failure\\ rate (\%)\end{tabular}  
         & \begin{tabular}[c]{@{}c@{}}Collision\\ rate (\%)\end{tabular} 
         & \begin{tabular}[c]{@{}c@{}}Off-road\\ rate (\%)\end{tabular} \\
\Xhline{2\arrayrulewidth}
BC + Route & 59.07{\scriptsize $\pm$1.45} & 8.66{\scriptsize $\pm$0.83} & 32.29{\scriptsize $\pm$1.38} & 6.63{\scriptsize $\pm$0.73} & 72.38{\scriptsize $\pm$1.32} & 9.02{\scriptsize $\pm$0.84} & 17.22{\scriptsize $\pm$1.11} & 6.69{\scriptsize $\pm$0.74}\\
MGAIL + Route & 77.77{\scriptsize $\pm$1.23} & \bf 7.24{\scriptsize $\pm$0.77} & 15.70{\scriptsize $\pm$1.08} & 3.03{\scriptsize $\pm$0.51} & 83.38{\scriptsize $\pm$1.10} & \bf 6.77{\scriptsize $\pm$0.74} & 9.48{\scriptsize $\pm$0.87} & 2.71{\scriptsize $\pm$0.48}\\
MGAIL + BC + Route & \bf 78.05{\scriptsize $\pm$1.22} & 8.95{\scriptsize $\pm$0.84} & \bf 14.09{\scriptsize $\pm$1.03} & \bf 2.23{\scriptsize $\pm$0.44} & \bf 85.69{\scriptsize $\pm$1.03} & 7.04{\scriptsize $\pm$0.75} & \bf 7.27{\scriptsize $\pm$0.77} & \bf 2.20{\scriptsize $\pm$0.43}\\
\Xhline{2\arrayrulewidth}
\end{tabular}}
\end{table*}

\subsubsection{\textbf{What is the performance on challenging scenarios?}}

To better understand how the model performs in more challenging scenarios, we evaluate it on test subsets of varying difficulty levels.
We first compute a score for each segment in the test set using the challenge classifier (section \ref{sec:risk_classifier}).
We then create four subsets of increasing levels of predicted collisions and close-calls (number of segments for logged and novel routes in parentheses): \emph{Low} (8,436 and 2,442), \emph{All} (82,198 and 18,593), \emph{Medium} (7,148 and 954), and \emph{High} (2,401 and 302), where \emph{All} refers to the entire \emph{Unbiased Test Set} and \emph{Route Generalization Test Set} for logged and novel routes.

\begin{figure}[th]
\vspace{-2ex}
\centering
\includegraphics[width=1.1\columnwidth]{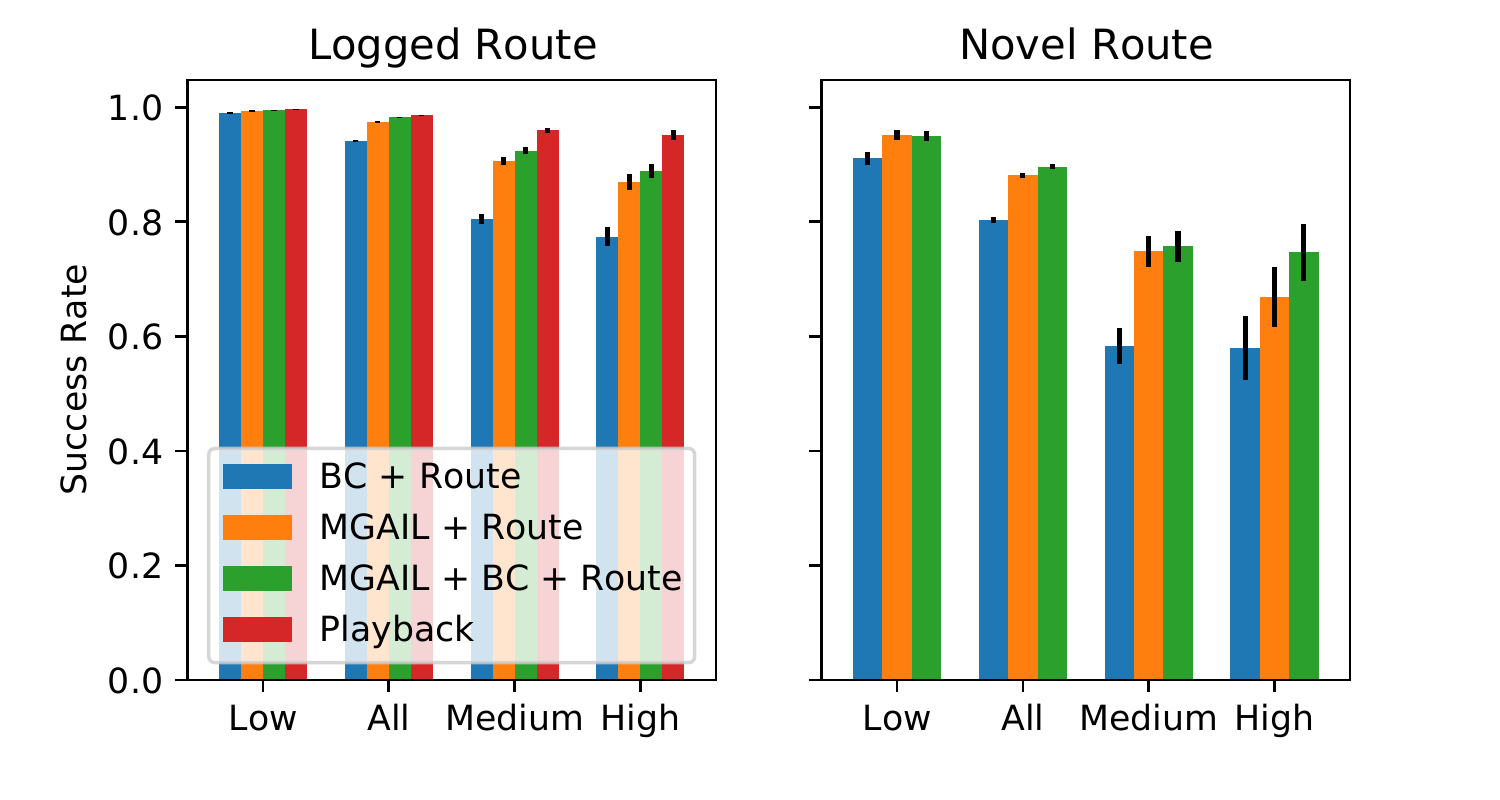}
\caption{Logged Route vs.\ Novel Route success rate for different challenge levels.}
\label{fig:risk_log_vs_ood}
\vspace{-2ex}
\end{figure}

We evaluate the model variants on each of these subsets, both for logged routes and novel routes.
As shown in Figure \ref{fig:risk_log_vs_ood}, the overall success rate of the expert and all model variants is lower on more challenging subsets, and the model performs worse on novel routes than on logged routes, as expected.
Variants that use MGAIL outperform BC across all challenge levels, but most acutely on the more difficult subsets, highlighting the importance of MGAIL in providing robustness to harder driving situations.
While MGAIL alone performs comparably to, if not better than, MGAIL + BC on the least challenging subset, MGAIL + BC achieves a higher success rate on the average and more difficult subsets for both logged and novel routes.
We observe the same patterns in the individual metrics that make up the overall success rate.

% \textit{What is the optimal mixture of losses and impact of closed-loop training?}
\subsubsection{\textbf{What is the impact of closed-loop training and how should MGAIL be prioritized vs.\ BC?}}

\begin{figure}[ht]
\vspace{-2ex}
\centering
\includegraphics[width=0.95\columnwidth]{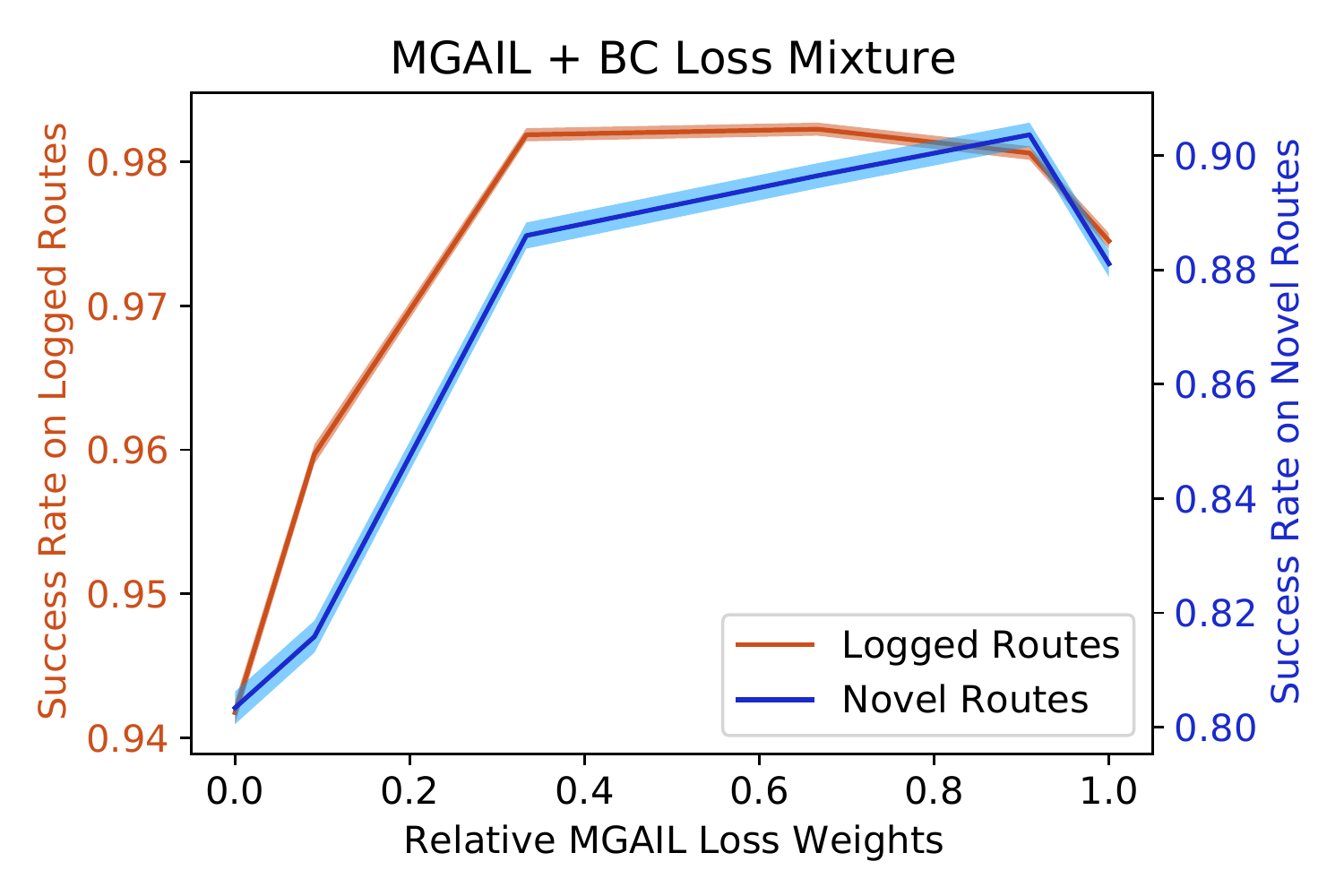}
\caption{Success rates of route-conditioned models on logged and novel routes for different relative MGAIL loss weights.}
\label{fig:bc_mgail_mixing}
\vspace{-2ex}
\end{figure}

Our results indicate that employing closed-loop training with MGAIL and adding the BC loss produces the highest success rates on both logged and novel routes. %
This leads to the question of which combination of adversarial and BC losses performs best. %
To determine this, we evaluate different MGAIL and BC policy loss weight mixtures and compare their success rates. %
In particular, we show success rates for different relative MGAIL loss weights (Figure \ref{fig:bc_mgail_mixing}), which set the ratio of the MGAIL policy loss weight ($\mathcal{\lambda_{P}}$) to the total policy loss weight ($\mathcal{\lambda_{P}} + \mathcal{\lambda_{BC}}$). %
% Figure \ref{fig:bc_mgail_mixing} shows the success rates for different loss mixtures. %
We obtain the best results by weighting the MGAIL loss more than the BC loss. %
Furthermore, for novel routes compared to logged routes, a higher relative MGAIL loss weight yields the highest success rate, providing additional evidence that MGAIL is beneficial for generalization. 

%\[
%    \text{Relative MGAIL loss weight}= \frac{ \text{MGAIL loss weight}}{\text{MGAIL loss weight} + \text{BC loss weight}}
%\]
%
% Our results in Table \ref{tab:2} show that among the route conditioned models BC exhibits the lowest route failure rate 
% BC + route has lowest route failure for logged routes but highest route failure for novel routes
% MGAIL + route has lowest route failure rate on novel routes but highest (equal to MGAIL + BC) on logged routes
%
%
%Commented out temporarily to see if we can move this figure next to table 1 to save space

%\textit{Can we use simpler observation encoders?}
%\textit{What is the impact of different route features?}

%\section{example figures and tables}
%Positioning Figures and Tables: Place figures and tables at the top and bottom of columns. Avoid placing them in the middle of columns. Large figures %and tables may span across both columns. Figure captions should be below the figures; table heads should appear above the tables. Insert figures and %tables after they are cited in the text. Use the abbreviation Fig. 1, even at the beginning of a sentence.

% \begin{figure}[thpb]
%   \centering
%   \framebox{\parbox{3in}{We suggest that you use a text box to insert a graphic (which is ideally a 300 dpi TIFF or EPS file, with all fonts embedded) because, in an document, this method is somewhat more stable than directly inserting a picture.}}
%   %\includegraphics[scale=1.0]{figurefile}
%   \caption{Inductance of oscillation winding on amorphous
%   magnetic core versus DC bias magnetic field}
%   \label{figurelabel}
% \end{figure}

\section{CONCLUSION AND FUTURE WORK}

We demonstrated a hierarchical model-based generative adversarial imitation learning (MGAIL) method that performs similarly to an expert demonstrator on a large unbiased sample of urban driving on key planning metrics.
We highlighted the importance of closed-loop training with MGAIL, as well as closed-loop evaluation with interactive agents in order to more accurately assess the planning agent's ability to generalize.
Although aggregate performance is similar to the expert, we showed that there still remains a gap on generalization to novel routes, as well as performance on challenging and rare scenarios.
This suggests further improvements must focus on this ``long-tail" of performance.
Refinement through reinforcement learning and reward shaping with safety constraints, or active curriculum learning on specific data subsets are two promising directions for future work.
In addition, training the planning agent directly alongside pre-trained interactive agents may improve the policy's ability to interact with and influence other actors, facilitating better generalization to novel goals.
Incorporating this policy into a complete AV platform to drive real-world vehicles is a natural next step, which will likely necessitate dynamic rerouting and adapting our action model to include realistic vehicle dynamics.
                                  % This command serves to balance the column lengths
                                  % on the last page of the document manually. It shortens
                                  % the textheight of the last page by a suitable amount.
                                  % This command does not take effect until the next page
                                  % so it should come on the page before the last. Make
                                  % sure that you do not shorten the textheight too much.
% Commented out in order to prevent the bibliography from being split with a
% page break. See
% https://tex.stackexchange.com/questions/113575/avoiding-page-break-in-the-latex-bibliography-list
% \addtolength{\textheight}{-12cm}

%%%%%%%%%%%%%%%%%%%%%%%%%%%%%%%%%%%%%%%%%%%%%%%%%%%%%%%%%%%%%%%%%%%%%%%%%%%%%%%%

%%%%%%%%%%%%%%%%%%%%%%%%%%%%%%%%%%%%%%%%%%%%%%%%%%%%%%%%%%%%%%%%%%%%%%%%%%%%%%%%

%%%%%%%%%%%%%%%%%%%%%%%%%%%%%%%%%%%%%%%%%%%%%%%%%%%%%%%%%%%%%%%%%%%%%%%%%%%%%%%%
%\section*{APPENDIX}
%Appendixes should appear before the acknowledgment.

\section*{ACKNOWLEDGMENT}
We thank Rami Al-Rfou, Ury Zhilinsky, Ben Sapp, James Philbin, and Jonathan Bingham for their helpful comments, and Aleksei Timofeev for his contributions to model-based imitation learning during his employment at Waymo.

\bibliography{iros-22-planning_agent}
\bibliographystyle{IEEEtran}

\end{document}